# The Segmentation Fusion Method On10 Multi-Sensors

Firouz Abdullah Al-Wassai[1], Dr. N.V. Kalyankar[2]
[1]Ph. D. Scholar, Computer Science Dept. (SRTMU), Nanded, India
[2]Principal, Yeshwant Mahavidyala College, Nanded, India
fairozwaseai@yahoo.com, drkalyankarnv@yahoo.com

*Abstract*: **The most significant problem may be undesirable effects for the spectral signatures of fused images as well as the benefits of using fused images mostly compared to their source images were acquired at the same time by one sensor. They may or may not be suitable for the fusion of other images. It becomes therefore increasingly important to investigate techniques that allow multi-sensor, multi-date image fusion to make final conclusions can be drawn on the most suitable method of fusion. So, In this study we present a new method Segmentation Fusion method (SF) for remotely sensed images is presented by considering the physical characteristics of sensors, which uses a feature level processing paradigm. In a particularly, attempts to test the proposed method performance on 10 multi-sensor images and comparing it with different fusion techniques for estimating the quality and degree of information improvement quantitatively by using various spatial and spectral metrics.**

## I. INTRODUCTION

Most of the newest remote sensing images provide data at different spatial, temporal, radiometric and Spectral resolutions, such as Landsat, Spot, Ikonos, Quickbird, Formosat, GeoEye or Orbview provide panchromatic PAN images at a higher spatial resolution than in their multispectral mode MS. Imaging systems somehow offer a tradeoff between high spatial and high spectral resolution, whereas no single system offers both [1]. This becomes remains challenging due to many causes, such as the various requirements, the complexity of the landscape, the temporal and spectral variations within the input data set [2]. So that image fusion has become a powerful solution to provide an image containing the spectral content of the original MS images with enhanced spatial resolution [3]. Often Image fusion techniques divided into three levels for processing fusion, namely: pixel level, feature level and decision level of representation. A large number of fusion or pan-sharpening techniques have been suggested to combine MS and PAN images with the promise to minimize color distortion while retaining the spatial improvement of the standard data fusion algorithms. For example the image fusion techniques based on pixel level in the lectures Summarized as the fallowing: **a)** Arithmetic Combination techniques: such as Bovey Transform (BT) [4-8]; Color Normalized Transformation (CN) [9, 10]; Multiplicative Method (MLT) [11]. **b)** Frequency Filtering Methods: such as High-Pass Filter Additive Method (HPFA) [12, 13], High –Frequency-Addition Method (HFA)[13] , High Frequency Modulation Method (HFM) [13] and The Wavelet transform-based fusion method (WT) [15-17]. **c)** Component Substitution fusion techniques: intensity–hue–saturation (HIS)*, YIQ and Hue Saturation Value (HSV) [18-22]. **d)** Statistical Methods: such as Local Mean Matching (LMM), Local Mean and Variance Matching (LMVM) [22-23], Regression variable substitution (RVS) [25-26], and Local Correlation Modeling (LCM) [27].

From previous studies to examine the benefits of image fusion techniques compared to their source images some them were acquired at the same time by one sensor (single-sensor, single-date fusion), for specific tasks yielded mixed results as well as contradiction results for same method for instance [19,28,29]. Thus, there is a need to investigate techniques that allow multi-sensor, multi-date image fusion is necessary to make final conclusions can be drawn on the most suitable method of fusion.

Therefore, this study, proposed a new method (SF) which uses a feature level processing paradigm for merging the images with different information i.e., spatial; spectral; temporal and radiometric resolution acquired by multi-sensors as well as comparing results it with selected methods from the mentions above, which that methods tested in our previous studies [30-33] as follows: HFA; HFM; RVS; IHS, HSV and Edge Fusion (EF) were much better than the others methods. This paper also devotes to concentrate on the analytical techniques for evaluating the quality of image fusion (F) by using various spatial and spectral metrics.

The paper organized as follows: Section II illustrates a new proposed scheme of SF method. Section III includes the quality of evaluation of the fused images; section IV covers the experimental results and analysis then subsequently followed by the conclusion in Section V.

## II. A NEW PROPOSED FUSION TECHNIQUE (SF)

Segmentation refers to the process of partitioning an image into multiple segments. The SF was developed specifically for a spectral characteristics preserving image merge. It is based on IHS transform coupled with a spatial domain filtering for feature extraction.

The principal idea behind a spectral characteristics preserving image fusion is that the high R of PAN image has to sharpen the MS image without adding new gray level information to its spectral components. An ideal fusion algorithm would enhance high frequency changes by Feature extraction such as edges and high frequency gray level changes in an image without altering the MS components in homogeneous regions. To facilitate these demands, two prerequisites: 1) color and spatial information have to be separated. 2) The spatial information content has to be segmented and manipulated in a way that allows adaptive enhancement of the images. The intensity $I_{LPF}$ of MS image is filtered with a low pass filter (LPF) [34] whereas the PAN image is filtered with an opposite high pass filter (HPF) [29, 35]. Basically HPF consists of an addition of spatial details, taken from the PAN into MS image. In this study, to extract the PAN channel high $P_{HPF}$ frequencies; a degraded or low-pass-filtered version of the PAN channel has to be created by applying the filter weights in a 3 x 3 convolution filter to computing a local average around each pixel in the image, is achieved. Since the goal of contrast enhancement is to increase the visibility of small detail in an image, subsequently, the (HPF) extracts the high frequencies using a subtraction procedure .This approach is known as Un-sharp masking (USM) [36]:

$$P_{USM} = P - P_{LPF} \quad (1)$$

When this technique is applied, it leads to the enhancement of all high spatial frequency detail in an image including edges, line and points of high gradient [37].

$$I^* = I_{LPF} + P_{UMS} \quad (2)$$

The low pass filtered intensity ($I_{LPF}$) of MS and the high pass filtered PAN band ($P_{UMS}$) are added and matched to the original intensity using the mean and standard deviation adjustment, which is also called adaptive contrast enhancement in [38], as the following:

$$I^*_{New} = \bar{I} + (I^* - \bar{I}^*)\frac{\sigma_I}{\sigma_{I^*}} \quad (3)$$

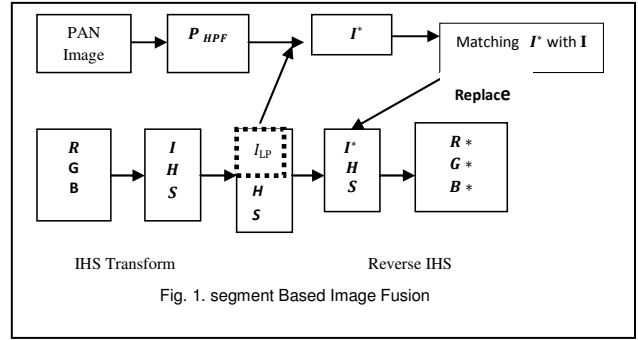

Fig. 1. segment Based Image Fusion

σ and Mean adaptation are, in addition, a useful means of obtaining images of the same bit format (e.g., 8-bit) as the original MS image [39]. After filtering and matching processing, the images are transformed back into the spatial domain with an inverse IHS and added together ($I^*$) to form a fused intensity component with the low frequency information from the low resolution MS image and the high-frequency information from the PAN image. This new intensity component and the original hue and saturation components of the MS image form a new IHS image. As the last step, an inverse IHS transformation produces a fused RGB image that contains the spatial resolution of the PAN image and the spectral characteristics of the MS image. An overview flowchart of the SF method is presented in Fig. 1

## III. QUALITY EVALUATION OF THE FUSED IMAGES

This section describes the various spatial and spectral quality metrics used to evaluate them for fused images. To explain the algorithms through this study, the pixel should have the same spatial resolution from two different sources that are manipulated to obtain the resultant image. Therefore, the MS images resample to the same size of PAN by neighbor manipulation. The spectral fidelity of the fused images with respect to the spectral characteristics of re-sampled original MS images while the spatial properties compared to the PAN image. The following notations will be used: P as a digital number DN for PAN image, $F_k$, $M_k$ are the measurements of each the brightness pixels values of the result image and the original MS image of band $k$, $\bar{M}_k$ and $\bar{F}_k$ are the mean brightness values of both images and are of size $M * N$, $Bv$ is the brightness value of image data $\bar{M}_k$ and $\bar{F}_k$.

A. **Spectral Quality Metrics**

1. **Deviation Index (DI)**[36]:
$$DI_k = \frac{1}{nm}\sum_{i}^{n}\sum_{j}^{m}\frac{|F_k(i,j)-M_k(i,j)|}{M_k(i,j)} \quad (4)$$

2. **Signal-to Noise Ratio (SNR)**[14]:
$$SNR_k = \sqrt{\frac{\sum_{i}^{n}\sum_{j}^{m}(F_k(i,j))^2}{\sum_{i}^{n}\sum_{j}^{m}(F_k(i,j)-M_k(i,j))^2}} \quad (5)$$

3. **Normalization Root Mean Square Error (NRMSE)** [30]:
$$NRMSE_k = \sqrt{\frac{1}{nm*255^2}\sum_{i}^{n}\sum_{j}^{m}(F_k(i,j)-M_k(i,j))^2} \quad (6)$$

B. *Spatial improvement evaluation*

1) **Filtered Correlation Coefficients (FCC):** In [31] this approach was introduced for the calculation of the FCC. A high-pass HP filter with a 3x3 Laplacian kernel is first applied to the PAN image and to each band of the fused image. Then the correlation coefficients between the HP filtered bands and the HP filtered PAN image are calculated. According to [1]. The FCC value close to one indicates high spatial quality.

2) **High Pass Deviation Index (HPDI):** This study employed the proposed quality metric in [40] to measure the amount of edge information from the PAN image is transferred into the fused images. This approach also, HP filter with a 3x3 Laplacian kernel applied to the PAN image $P$ and the fused image $F_{HP\,k}$. Then the deviation index between the $P_{HP}$ and the fused $F_{HP\,k}$ images would indicate how much spatial information from the PAN image has been incorporated into the *MS* image to obtain HPDI as follows [40]:

$$HPDI_k = \frac{1}{nm}\sum_{i}^{n}\sum_{j}^{m}\frac{|F_{HP\,k}(i,j)-P_{HP}(i,j)|}{P(i,j)} \quad (7)$$

The larger value HPDI the better image quality. Indicates that the fusion result it has a high spatial resolution quality of the image.

3) **Contrast Statistical Analysis (CSA)*:*** many formulize were found in lecture to calculate the contrast as Modulation transfer function (MTF) in [31] and referred to Michelson Contrast $C_M$ in [33]. This study used the CSA to evaluation the quality of the spatial resolution based on contrast calculation of each the edge and homogenous regions in [41].

IV. EXPERIMENTAL &ANALYSIS RESULTS

This work is an attempt to study the quality of the images fused for multi-sensor and multi-data with various characteristics. The above assessment techniques are tested results for different image fusion techniques including: EF, HFA, HFM, HSV, IHS, RVS, and SF methods. The pairs of images were geometrically registered to each other. The original MS& Pan images are shown in (Fig.2) and the fused images are shown in (Fig. 2) as well as the data of the Image Sources and the sensor characteristics tabulated in table (1). Here, to explain the results well are denoted for each pairs as Sen.1, 2, 3…, 10 in table (1).

Table (1): Test Image Sources, Location and Imaging Sensor Characteristics

| Test case No | Test Image Pairs Sources | ground resolution(m) | Geographical Location | Spectral range of Pan (μm) | Spectral range of Multispectral (μm) |
|---|---|---|---|---|---|
| Sen.1 | ASTER MS&IRS-1C PAN3 | (15,5) |  | 0.51 – 0.73 | VNIR (0.52-0.60 )<br>(0.63-0.69)<br>(0.78-0.86 ) |
| Sen.2 | IKONOS-2 MS&PAN | (4,1) | part of Sherbrooke city area, Quebec, Canada, | 760 - 850 | B(455 – 520)<br>G(510 - 600 )<br>R(630 - 700 |
| Sen.3 | SPOT5 MS& IKONOS PAN | (20,1) | near Santo Domingo de la Calzada, Spain | 0.45 – 0.90 | G(0.50 - 0.59)<br>R(0.61 - 0.68)<br>NIR (0.79-0.89) |
| Sen.4 | LANDSAT TM & SPOT PAN | (30,10) | Ningxia area, the western of China | 0.51 – 0.73 | B(0.45 - 0.52)<br>G(0.52 - 0.60)<br>R(0.63 - 0.69) |
| Sen.5 | SPOT HRV MS & PAN | (20,10) | Tang – Dynasty in the PR China | 0.51 – 0.73 | G(0.50 - 0.59)<br>R(0.61 - 0.68)<br>NIR(0.79 - 0.89) |
| Sen.6 | LANDSAT TM MS & SPOT5 PAN | (30,10) | Tang – Dynasty in the PR China | 0.51 – 0.73 | B(0.45 - 0.52)<br>G(0.52 - 0.60)<br>R(0.63 - 0.69) |
| Sen.7 | LANDSAT TM MS & IRS-1C PAN | ([30,5.8) | Tang – Dynasty in the PR China | 0.51 – 0.73 | B(0.45 - 0.52)<br>G(0.52 - 0.60)<br>R(0.63 - 0.69) |

| Sen.8 | IRS -1C III & PAN | (23.5,5.8) | Tang – Dynasty in the PR China | 0.51 – 0.73 | G(0.52 - 0.59) R(0.62 - 0.68) NIR(0.77 - 0.86) |
|---|---|---|---|---|---|
| Sen.9 | SPOT4 MS& SAR | | plantation Au-Ku in western of Taiwan. | | G(0.52 - 0.59) R(0.62 - 0.68) NIR(0.77 - 0.86) |
| Sen.10 | Quickbird MS&PAN | (2.8, 0.7) | Pyramid area of Egypt | 0.45 – 0.90 | B(0.45 - 0.52) G(0.52 - 0.60) R(0.63 - 0.69) |

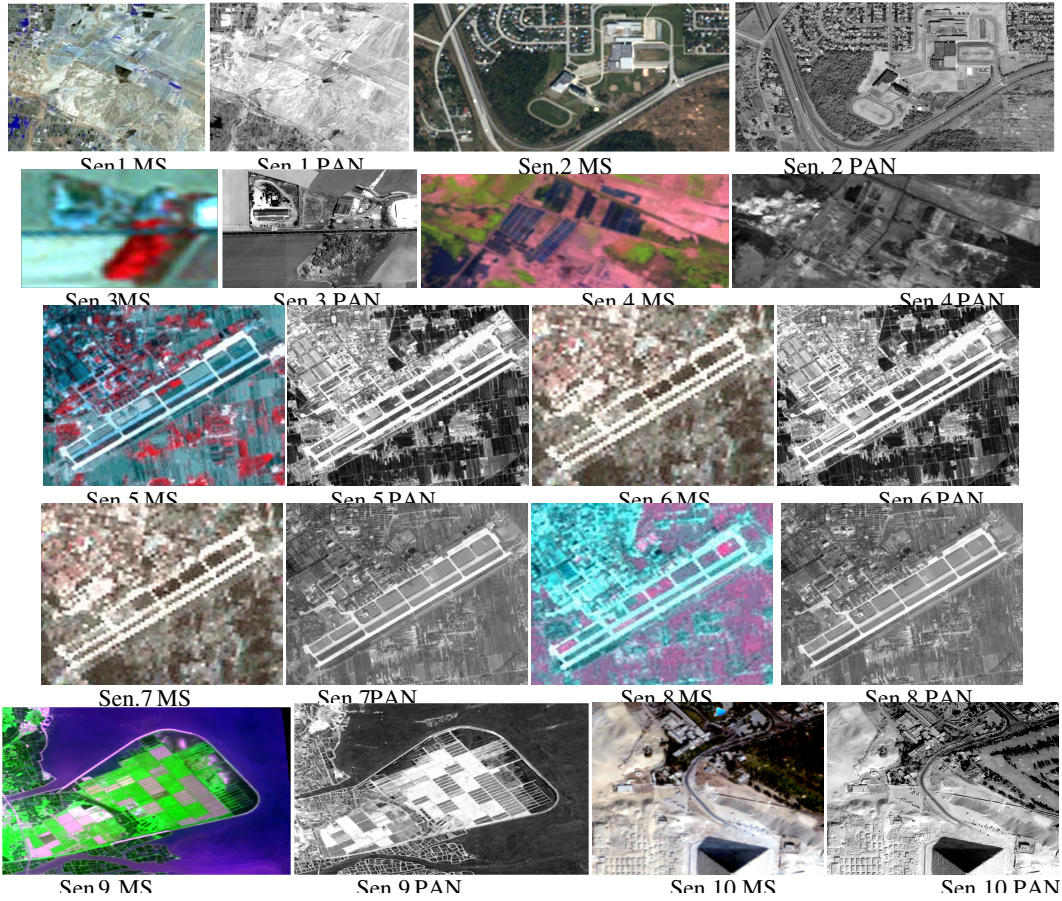

Fig.1: The Original MS & PAN Images for Fusion

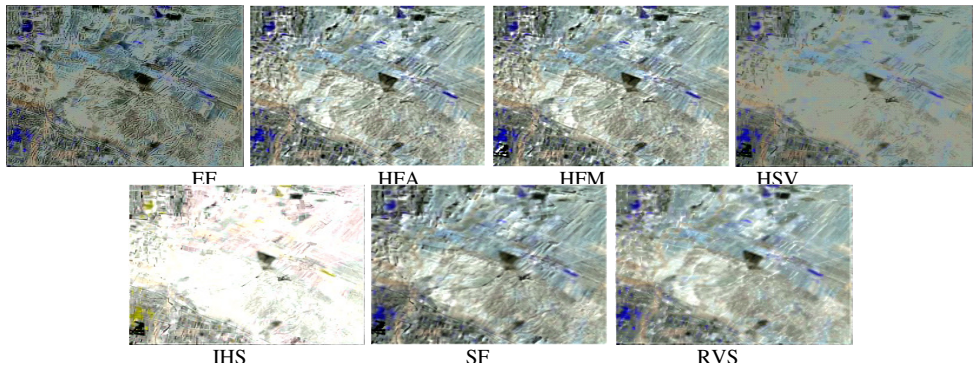

Fig.2: A set of 7 Fused Images From Sen. 1 In Fig.1

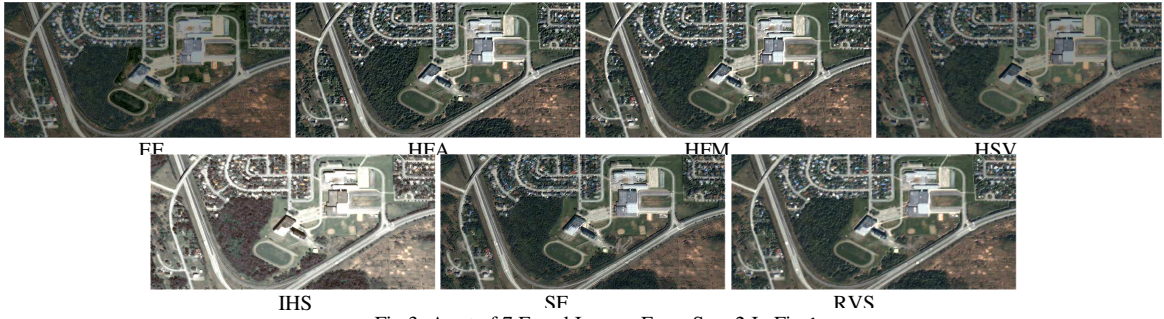
Fig.3: A set of 7 Fused Images From Sen. 2 In Fig.1

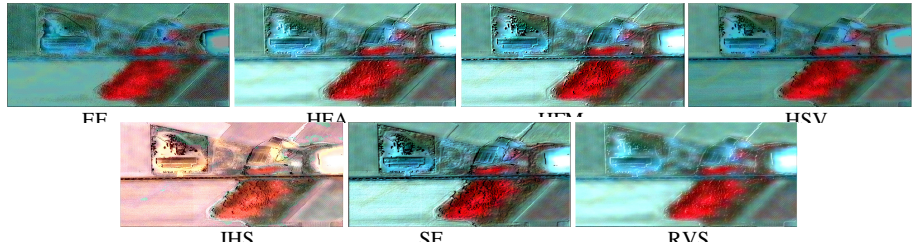
Fig.4: A set of 7 Fused Images From Sen. 3 In Fig.1

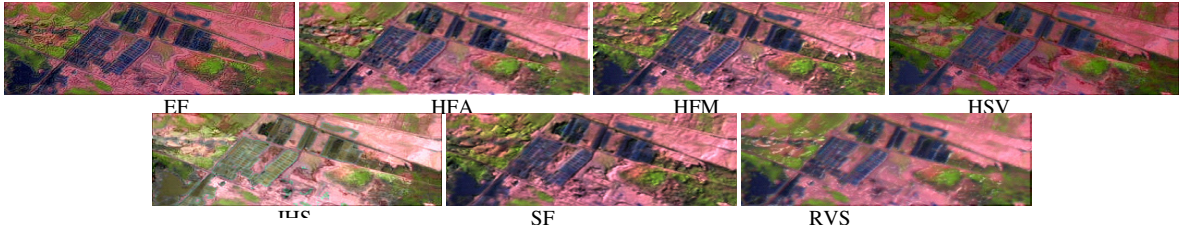
Fig.5: A set of 7 Fused Images From Sen. 4 In Fig.1

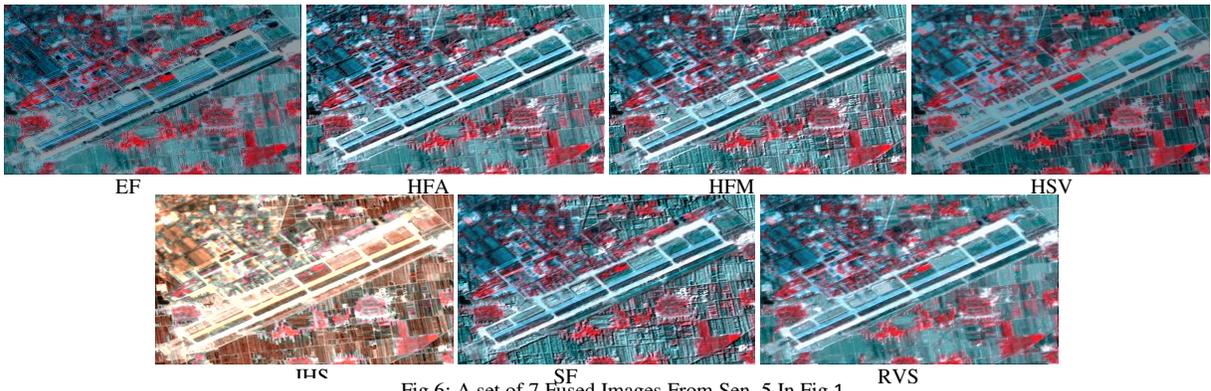
Fig.6: A set of 7 Fused Images From Sen. 5 In Fig.1

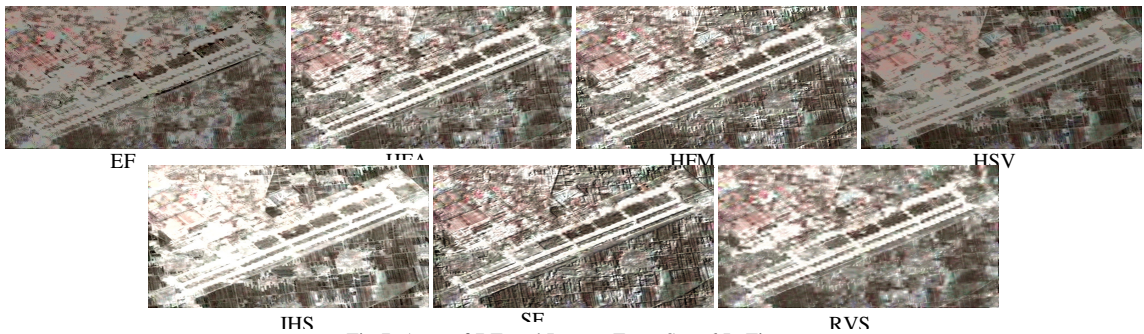
Fig.7: A set of 7 Fused Images From Sen. 6 In Fig.1

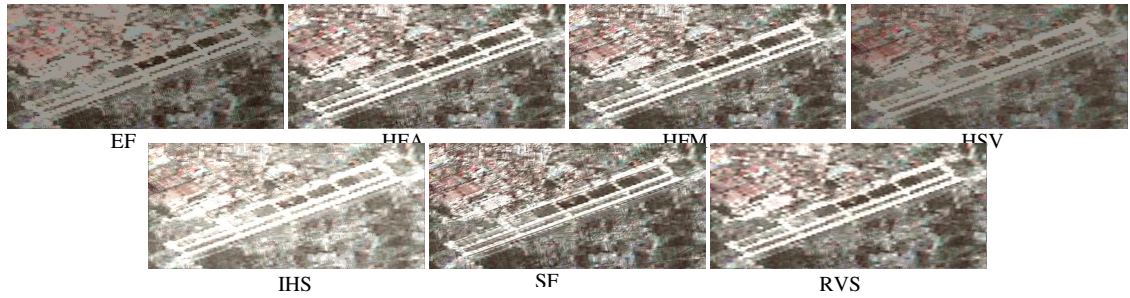
Fig.8: A set of 7 Fused Images From Sen. 7 In Fig.1

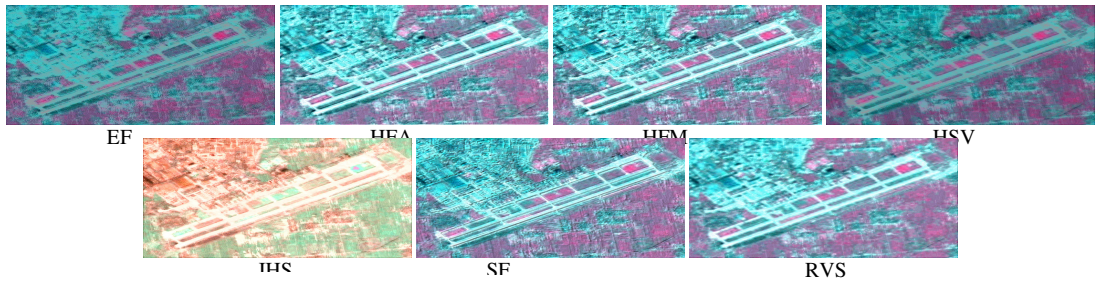
Fig.9: A set of 7 Fused Images From Sen. 8 In Fig.1

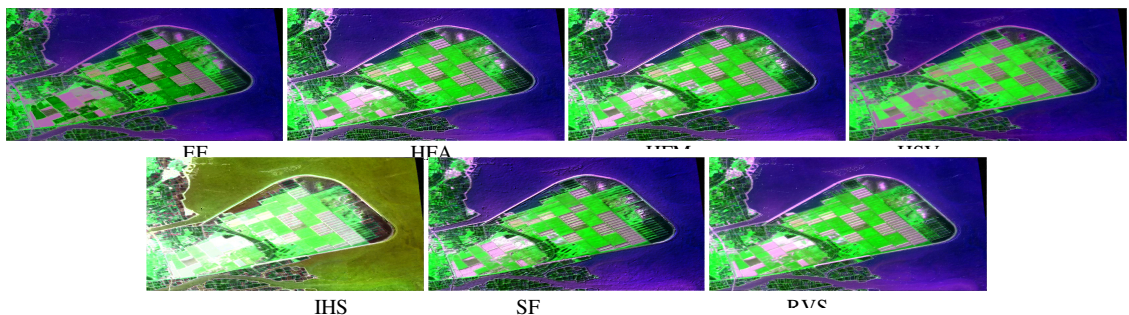
Fig.10: A set of 7 Fused Images From Sen. 9 In Fig.1

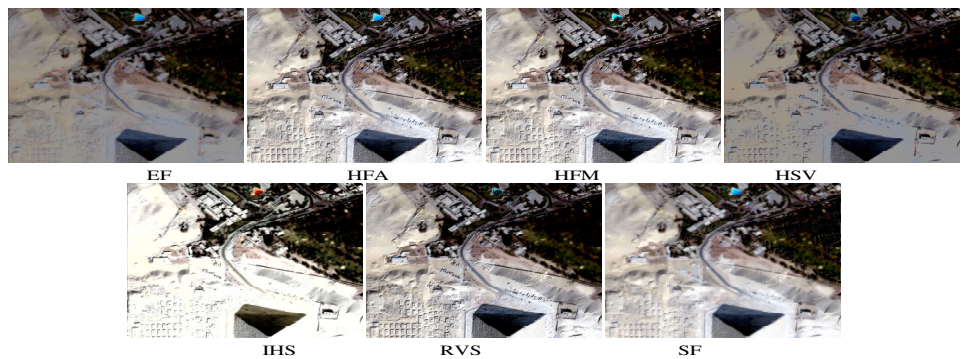
Fig.11. A set of 7 fused images from Sen.10 in Fig. 1.

### A. ANALYSISES RESULTS
#### a. *Spectral Quality Metrics Results:*

Results of the SNR, NRMSE and DI appear changing significantly. It can be observed from the diagram Fig. 13 for results SNR, NRMSE & DI of the fused image for each pairs images, the proposed SF method gives the best results with respect to the other methods for all sensors excepted Sen.10. Means that this method maintains most of information spectral content of the original MS data set which gets the same values presented the lowest value of the NRMSE and DI as well as the high of the SNR. Also, the IHS,

HSV and EF methods have the lowest values of SNR with high values for NRMSE and DI.

Due to the large values DI of IHS in the Fig.12c totally veiled Note the differences are obvious to the rest of the modalities of the merger has been canceled so the values of DI for IHS in the Fig.12d. We note from Fig.12d the differences are clear to the values of DI for the various techniques applied to all images of the various sensors applied during this study. Less value for DI was with SF method except both Sen.9 & 10 and that their values are also high for the various technologies. And study the impact of the results of the different techniques of integration according to the different sensors in the maintenance of the spectral features of the original images and found that their performance is almost the same performance, whether bad or good. The best result of the various methods is a SF method. The best outcome of the consolidation of the various methods with the Indian Sen.8 of sensors, while the worst results for all techniques of integration on all image pairs of the different sensors with the Sen.9 & 10. Therefore, in the future need to study the methods meet the requirements of efficient integration of these sensors.

### b. *Spatial Quality Metrics Results:*

Fig. 13 show the result of different sensors fused images using various methods. It is clearly that the seven fusion methods are capable of improving the spatial resolution with respect to the original MS image for all different sensors. Note from Fig.13a, where disorder results cannot distinguish between which is better, or vice versa. The reason for that is the different spatial and spectral feature recorded for the various sensors. So is not recommended for using as the spatial criterion with images of different sensors. According to the computation results, FCC in Fig.13c the increase FCC indicates the amount of edge information from the PAN image transferred into the fused images in quantity of spatial resolution through the merging. The maximum results of FCC From Fig.13c were with all methods except the EF and RVS methods. Also, the lowest enhancements of the spatial resolution for all sensors were with Sen.9&10. The results of HPDI in Fig.13b better than

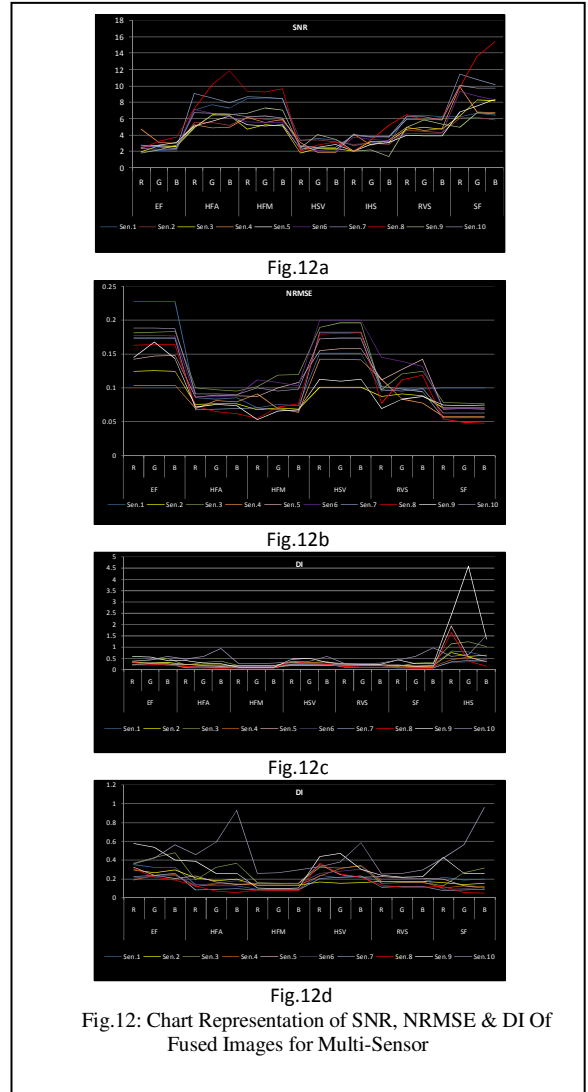

Fig.12a

Fig.12b

Fig.12c

Fig.12d

Fig.12: Chart Representation of SNR, NRMSE & DI Of Fused Images for Multi-Sensor

FCC or Contrast results in Fig.13 it is appear changing significantly. The approach of HPDI as the spatial quality metric is more important than the other spatial quality matrices to distinguish the best spatial enhancement through the merging. It can be observed that from Fig.13b the maximum results of HPDI it was with the EF and SF methods. The EF has the highest values of HPDI even so the details do not match the spatial image of the original for enhancing the spatial resolution because it depends on the emphasis filtering techniques.

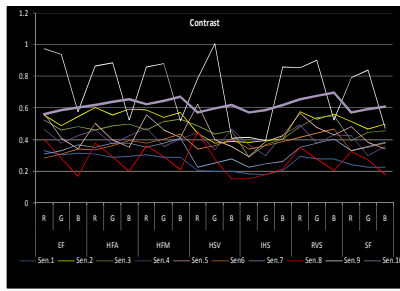 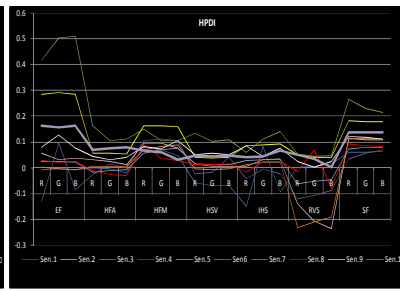 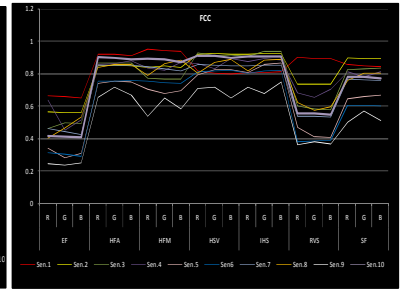

Fig.13a: Contrast   Fig.13b: HPDI   Fig.13c: FCC

Fig.13: Chart Representation of Contrast, HPDI & FCC of Different Sensors Fused Images

## V. CONCLUSION

In ideal condition, a good image fusion method tries to generate the image on any sensor would obtain if it had the same spectral signature of the original MS sensor with the spatial resolution of the PAN sensor. So, this paper goes through the comparative studies undertaken by different measures for assessing the quality of fused images has been conducted. A total of 10 image pairs with different types of land covers multi-sensor and multi-data are used to examination the proposed a approach SF method and compare it with Image Fusion techniques as follows: HFA, HFM, IHS, RVS, HSV and EF. Experimental results with spatial and spectral quality matrices evaluation further show that the proposed SF technique based on feature level fusion maintains the spectral integrity for MS image as well as improved as much as possible the spatial quality of the PAN image with all 10 image pairs except with Sen.10. The use of the SF based fusion technique is strongly recommended if the goal of the merging is to achieve the best representation of the spectral information of multispectral image and the spatial details of a high-resolution PAN image. Because it is based on Component Substitution fusion techniques coupled with a spatial domain filtering. It utilizes the statistical variable between the brightness values of the image bands to adjust the contribution of individual bands to the fusion results to reduce the color distortion.

Also, observed the impact of merge different sensor images during this study that the different methods of fusion techniques on images for different sensors are the same performance, whether good or vice versa, whether different sensors or images did not differ. In other words, the impact of the merger is no different in different images of different sensors appear almost the same result. It also concluded through this study that all fusion techniques applied during this study, the results were bad with the sen.10 and the reason is due to the large difference in the percentage of resolution of spatial between MS and PAN images. And it is proposed to study the modalities for the integration to meet that requirement to those for the integration of these sensors.

The HPDI gave the smallest different ratio between the image fusion methods, therefore, it is strongly recommended to use HPDI for measuring the spatial resolution because of its mathematical and more precision as quality indicator. As concluded in this study the accuracy of the different criteria used to assess the performance efficiency of the merger. Therefore we recommend in the future studying more thoroughly the different criteria for evaluating the performance of the merger.

Short Biodata of the Author

Firouz Abdullah Al-Wassai. Received the B.Sc. degree in physics from University of Sana'a, Yemen in 1993; the M. Sc. degree from Bagdad University, Iraq in 2003. Currently, she is Ph. D. scholar in computer Science at department of computer science (S.R.T.M.U), Nanded, India.

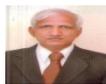 Dr. N.V. Kalyankar,He is a Principal of Yeshwant Mahvidyalaya, Nanded(India) completed M.Sc.(Physics) from Dr. B.A.M.U, Aurangabad. In 1980 he joined as a leturer in department of physics at Yeshwant Mahavidyalaya, Nanded. In 1984 he completed his DHE. He completed his Ph.D. from Dr.B.A.M.U, Aurangabad in 1995. From 2003 he is working as a Principal to till date in Yeshwant Mahavidyalaya, Nanded. He is also research guide for Physics and Computer Science in S.R.T.M.U, Nanded. 03 research students are successfully awarded Ph.D in Computer Science under his guidance. 12 research students are successfully awarded M.Phil in Computer Science under his guidance He is also worked on various boides in S.R.T.M.U, Nanded. He is also worked on various bodies is S.R.T.M.U, Nanded. He also published 34 research papers in various international/national journals. He is peer team member of NAAC (National Assessment and Accreditation Council, India). He published a book entitled "DBMS concepts and programming in Foxpro". He also get various educational wards in which "Best Principal" award from S.R.T.M.U, Nanded in 2009 and "Best Teacher" award from Govt. of Maharashtra, India in 2010. He is life member of Indian "Fellowship of Linnaean Society of London (F.L.S.)" on 11 National Congress, Kolkata (India). He is also honored with November 2009.